\documentclass[10pt,twocolumn,letterpaper]{article}

\usepackage{cvpr}
\usepackage{times}
\usepackage{epsfig}
\usepackage{graphicx}
\usepackage{mathtools,amsmath}
\usepackage{amssymb}

\usepackage{authblk}
\usepackage{comment}
\usepackage{multirow}
\usepackage{float}
\usepackage[caption = false]{subfig}
\usepackage{adjustbox}
\usepackage[none]{hyphenat}
\usepackage{array}
\newcolumntype{P}[1]{>{\centering\arraybackslash}p{#1}}

\usepackage[pagebackref=true,breaklinks=true,letterpaper=true,colorlinks,bookmarks=false]{hyperref}

\cvprfinalcopy 

\def\cvprPaperID{0001} 

\ifcvprfinal\fi
\begin{document}

\title{What am I Searching for: Zero-shot Target Identity Inference in Visual Search}

\author[1,2]{Mengmi Zhang}
\author[1,2]{Gabriel Kreiman}

\affil[ ]{\small \{mengmi.zhang@childrens, gabriel.kreiman@tch\}.harvard.edu\normalsize}
\affil[1]{Children's Hospital, Harvard Medical School}
\affil[2]{Center for Brains, Minds and Machines}

\maketitle

\begin{abstract}
   Can we infer intentions from a person's actions? As an example problem, here we consider how to decipher what a person is searching for by decoding their eye movement behavior. We conducted two psychophysics experiments where we monitored eye movements while subjects searched for a target object. We defined the fixations falling on  \textit{non-target} objects as ``error fixations". Using those error fixations, we developed a model (InferNet) to infer what the target was. InferNet uses a pre-trained convolutional neural network to extract features from the error fixations and computes a similarity map between the error fixations and all locations across the search image. The model consolidates the similarity maps across layers and integrates these maps across all error fixations. InferNet successfully identifies the subject's goal and outperforms competitive null models, even without any object-specific training on the inference task.
\end{abstract}

\vspace{-4mm}
\section{Introduction}

\begin{figure}[t]
\begin{center}
\includegraphics[width=7.5cm, height = 5.5cm]{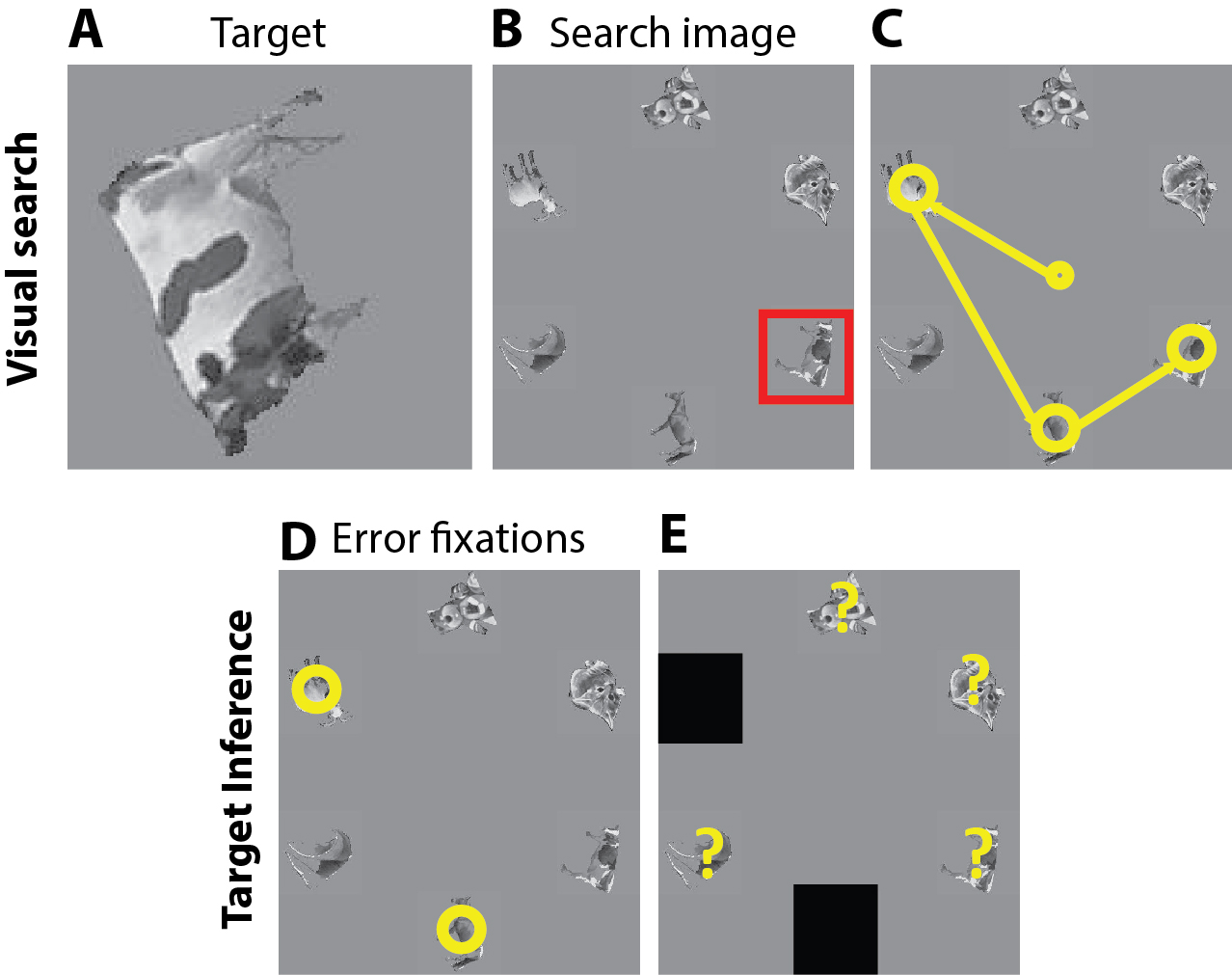}
\end{center}
   \caption{\textbf{Illustration of the target inference problem.} Human subjects were instructed to move their eyes to search for a given target (A) in the search image (B) irrespective of changes in size, rotation angles, or other format changes \cite{zhang2018finding}. The red box shows the  target location in the search image (not shown in the actual experiment). The visual search task resulted in a sequence of fixations (C, yellow circles).  In this example, the subject required 3 fixations to find the target. We defined the 2 fixations on \textit{non-target} objects as ``error fixations". In the target inference task, given the error fixations recorded from the psychophysics visual search task (D, yellow circles), the model infers what the subject was searching for (E, yellow question marks). \vspace{-5mm}
   }\vspace{-1mm}
\label{fig:intro}
\end{figure}

Inferring goals from actions is a central challenge for human-machine interactions and modeling human decisions. We consider the problem of inferring goals during eye movement behavior. Eye movements reflect rich information about the complex cognitive states of the brain, including thought processes and goals \cite{castelhano2009viewing,henderson2013predicting,betz2010investigating,iqbal2004using}. 

In a visual search task, ``error fixations'' prior to locating the target (\textbf{Fig.~\ref{fig:intro}}) are more likely to be on objects similar to the target \cite{eckstein2007similar,alexander2011visual}. Therefore, it is possible to decode target information from the eye movements \cite{borji2015eyes,zelinsky2013eye,rajashekar2006visual}.
Existing target decoding methods are limited in using elementary search statistics \cite{rajashekar2006visual}, or handcrafted features~\cite{borji2015eyes,zelinsky2013eye}. Moreover, existing approaches have only been tested with pre-defined classes and limited object set sizes. Those models do not generalize to infer targets from arbitrary classes. In contrast, here we propose a \emph{zero-shot model}, the Inference Network (InferNet). InferNet applies knowledge from object recognition to target inference. We tested InferNet on two visual search tasks \cite{zhang2018finding} and show that it can predict human goals \emph{without any re-training on new tasks}.

\section{InferNet}\label{sec:model}

\begin{figure*}[t]
\begin{center}
\includegraphics[width=15cm]{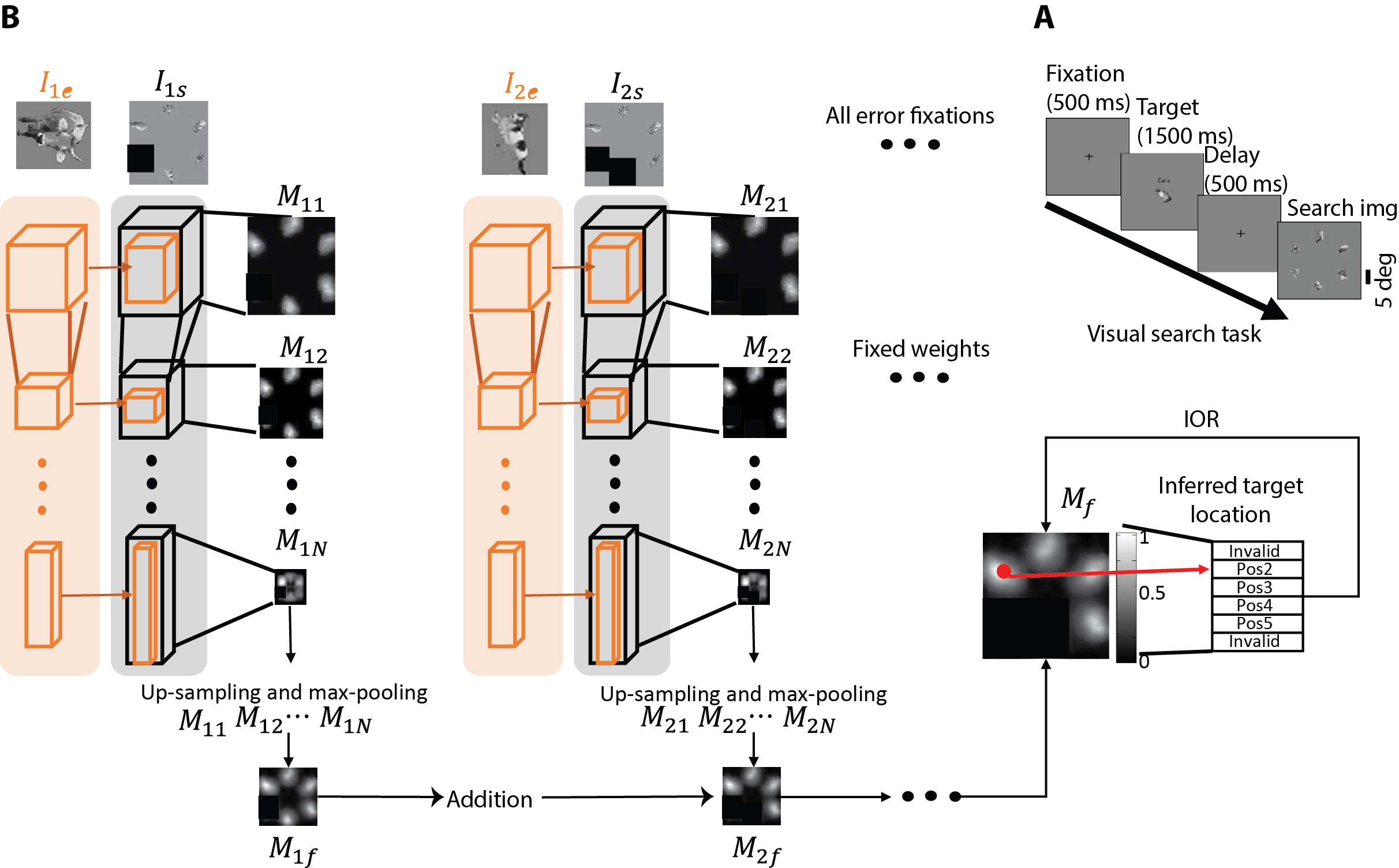}
\end{center}
   \caption{\textbf{Architecture of InferNet}. \textbf{A} Schematic of the visual search task \cite{zhang2018finding}. \textbf{B} InferNet takes two inputs: the object $I_{ie}$ at the error fixation $i$ and the search image $I_{is}$ with the object at the error fixation covered with a black mask. The error fixations are recorded from human subjects in the visual search task (\textbf{A}). See Sec.~\ref{sec:model} for model details.
}
\label{fig:model}
\end{figure*}


We provide an overview of 
the proposed zero-shot deep network (InferNet, \textbf{Fig.~\ref{fig:model}}).
Error fixations share more visual feature similarities with the target than with distractors \cite{eckstein2007similar,alexander2011visual,wolfe2007guided}.
Our model hypothesizes that locations showing higher feature similarity to error fixations are more likely to represent the subject's goal. Given $T$ error fixations with coordinates $(x_i,y_i)$ where $1\le i \le T$, the task is to  predict a probabilistic map $M_{f}$ of where the search target is most likely to be (\textbf{Fig.~\ref{fig:model}}). We take the maximum on $M_f$ as the current guess location. If the current guess location overlaps with the target object location, the inference is deemed successful. Otherwise, after each incorrect guess, the map is updated by removing the erroneous inference location on $M_f$ and the process continues iteratively.

The model uses a pre-trained deep convolutional network applied to the error fixations (\textbf{Prior Network (PN)}) and to the search image (\textbf{Likelihood Network (LN)}).
\textbf{PN} takes the cropped area $I_{ie}$ ($28 \times 28$ pixels) centered at error fixation $i$ as input and outputs feature maps across layers. We define $I_{is}$ as the search image, with the objects at past error fixations $1,...,i$ covered in black. \textbf{LN} modulates the feature maps from $I_{is}$, generating likelihood maps ($M_{i1}$, $M_{i2}$, $...$, $M_{ij}$, $...$,$M_{iN}$) where $j$ denotes the layer ($1 \le j \le N$). These maps are concatenated and max-pooled to produce the final likelihood map $M_{if}$ for error fixation $i$ which tracks the parts of the image that are most similar between $I_{ie}$ and $I_{is}$. InferNet adds the maps $M_{if}$ across all $T$ error fixations to build the final inference map $M_f$.    
\section{Experiments}\label{sec:exp}

We tested InferNet on data from two invariant visual search tasks using object arrays and natural images\cite{zhang2018finding}. We filtered out those fixations on targets, and only used error fixations (\textbf{Fig. \ref{fig:intro}}). The appearance of the target object in the search image was different from that in the target image.

We evaluate the model by the number of guesses required to infer the target as a function of the number of error fixations $T$. 
Because the target inference difficulty varies over trials, we report the relative performance $P_r$: $P_r(T) = \frac{A_c(T) - A_m(T)}{A_c(T)}$, where $A_m(T)$ is the average number of guesses required by InferNet and $A_c(T)$ is the average number of guesses required by chance. 
The larger $P_r(T)$, the more efficient the inference process is. 

We compared InferNet with alternative null models, including chance (random guess from the remaining possible locations), template matching (slide error fixation patterns over the search image), bottom-up saliency (Itti-Koch, \cite{itti1998model}), and RanWeight (InferNet with random weights without training). The null models proposed an inference map, and target selection procedure was the same as with InferNet. We made all the data (images, eye movements, and source code) publicly available online\footnote{https://github.com/kreimanlab/HumanIntentionInferenceZeroShot}.

\begin{figure}[t]
\begin{center}
\includegraphics[width=7.5cm]{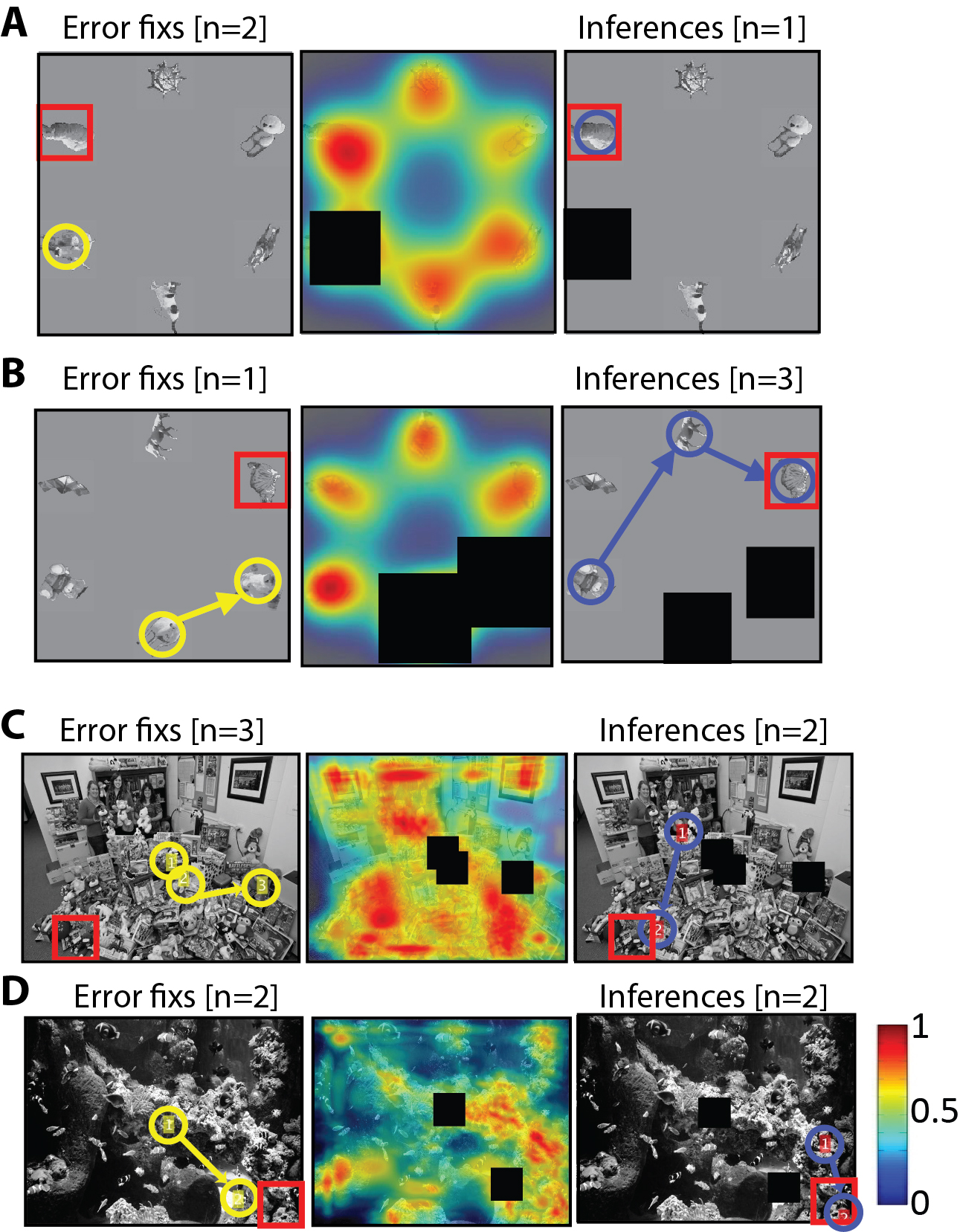}
\end{center}
   \caption{\textbf{Example target inference trials}. \textbf{A-B} Two object array trials. \textbf{C-D} Two natural image trials. Given the ``error fixations" (yellow circles, column 1), InferNet predicts the probabilistic map $M_f$ overlaid on the stimuli (Column 2, scale on the right). The red box (Column 1, 3) denotes the target location. The black boxes show the error fixation locations in $M_f$. The blue circles show the inferred locations (Column 3). 
\vspace{-5mm}}\vspace{-3mm}
\label{fig:qualatative}
\end{figure}

\begin{figure}[t]
\centering
\includegraphics[width=5.5cm]{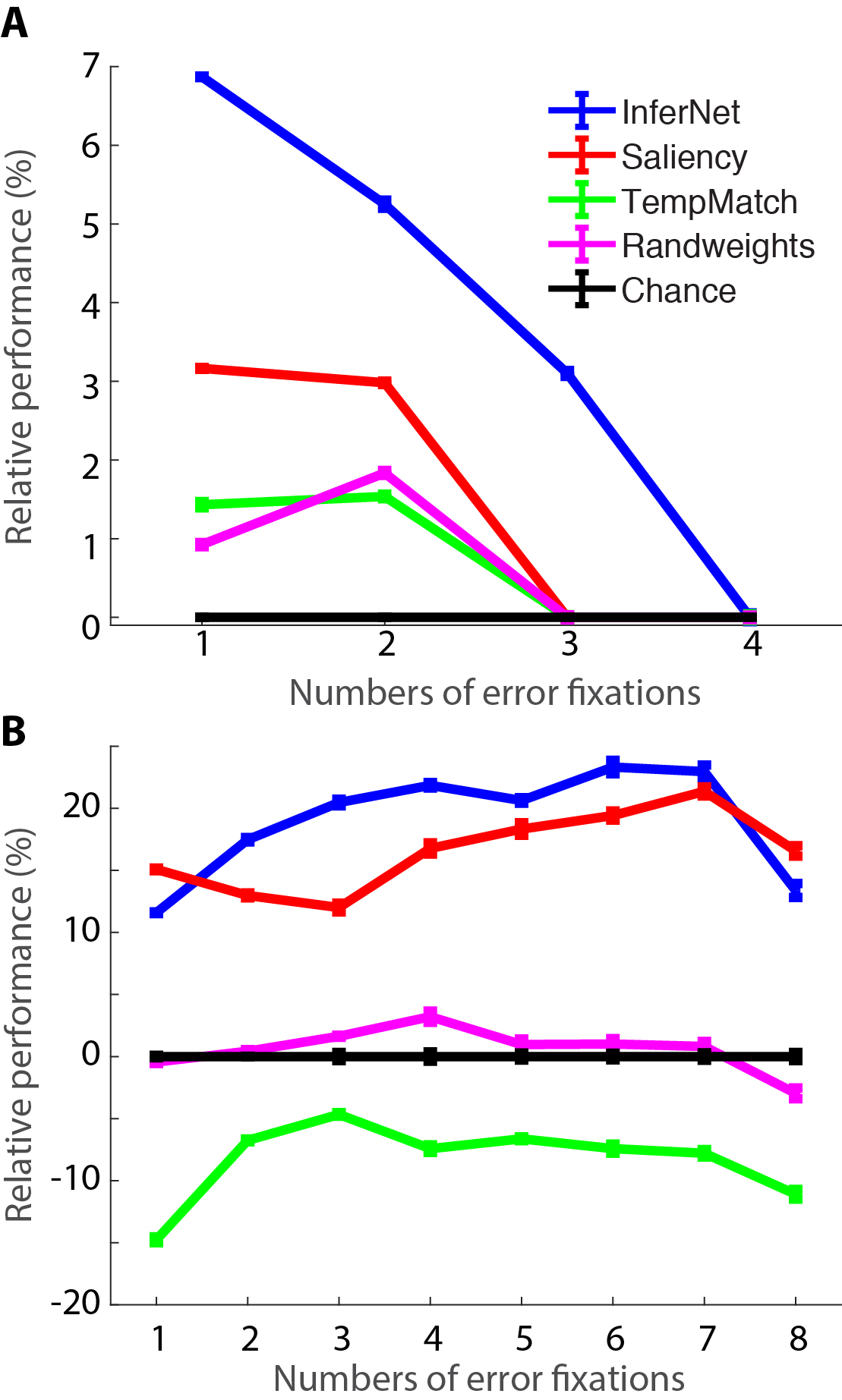}

\caption{\textbf{Evaluation of model inference performance for object arrays (A) and natural images (B)}. See Sec.~\ref{sec:exp}
    for descriptions of evaluation metrics and comparative models.
    Error bars are standard error of the mean. 
    \vspace{-3mm}}\label{fig:quantative}
\end{figure}

\subsection{Object arrays}

\textbf{Figure~\ref{fig:qualatative}} shows examples illustrating how the model efficiently inferred the target location given only 1-3 error fixations on object arrays and natural images. In \textbf{Fig.~\ref{fig:qualatative}A}, a subject made one error fixation on the cow (yellow circle) which looks similar to the target (red square) before finding the target sheep. Given this single error fixation, InferNet determined that the subject was probably looking for a sheep (blue circle) instead of the 4 remaining distractors. 

InferNet showed an overall improvement of $3.8\pm 3 \%$ with respect to the chance model over all error fixations (\textbf{Fig.~\ref{fig:quantative}A}, blue). Even with a single error fixation, InferNet could infer the target $6.9\%$ faster than chance. While random guessing would correctly land on the target within 3 guesses, InferNet only required 2.80 $\pm$ 0.01 guesses. 

None of the null models reached the performance level of InferNet (\textbf{Fig.~\ref{fig:quantative}A}, $P<10^{-19}$, two-tailed t-test), except with 4 error fixations where none of the models were above chance. Although we took precautions to normalize \emph{average} low-level features, InferNet can capitalize on shared low-level features between error fixations and the target. Performance for the bottom-up saliency model (IttiKoch) was better than the chance model but still below InferNet, suggesting that there is additional target information embedded in error fixations. The random weights (RanWeight) and pixel template matching (TempMatch) models showed minimal improvements compared to random locations (\textbf{Fig.~\ref{fig:quantative}A}), suggesting that the discriminative features learnt for image classification are important for target inference.




\subsection{Natural scenes}

The experiment reported so far focused on images consisting of segmented objects at discrete locations, presented on a uniform background, at fixed positions equidistant from the center of the image. In the real world, visual search takes place in cluttered environments involving non-segmented objects amidst a complex background. Next, we evaluated InferNet in the more challenging continuous domain of natural images (\textbf{Fig.~\ref{fig:qualatative}} and \textbf{Fig.~\ref{fig:quantative}B}).

\textbf{Figure~\ref{fig:qualatative}C} shows an example where InferNet successfully determined the subject's goal in natural images. The error fixations fell on plush toys, such as teddy bears. Based on the features of error fixations, the inference map showed high activations around plush toys regions. InterNet correctly inferred the target within 2 guesses. 

InferNet showed significant improvements of $19\pm 4\%$ compared to chance (\textbf{Fig.~\ref{fig:quantative}B}). InferNet outperformed all the alternative null models (\textbf{Fig.~\ref{fig:quantative}B}, $P< 10^{-26}$, two-tailed t-test). The bottom-up saliency model (IttiKoch) performed high among all the null models because target objects were  salient and relatively large. Template matching performed worse than the chance model. 



\begin{table}[t]
\scriptsize
\begin{center}

    \begin{tabular}{|c|c|c|c|c|c|c|c|c|}
      \hline
 \multirow{2}{*}{\shortstack{Error \\ Fixations}} & \multicolumn{8}{|c|}{Top $N$ category inference accuracy $\%$} \\ \cline{2-9}
         & 1    & 2    & 4    & 8    & 16   & 32   & 64   & 128  \\ \hline
       1 & 6 & 8 & 11 & 13 & 17 & 29 & 38 & 55 \\
       2 & 4 & 9 & 14 & 20 & 23 & 33 & 46 & 65 \\
       3 & 3 & 10 & 16 & 25 & 28 & 38 & 51 & 72 \\
       4 & 0    & 13 & 20 & 20 & 30 & 39 & 54 & 74 \\
       5 & 3 & 11 & 23 & 28 & 34 & 42 & 56 & 74 \\
       6 & 0    & 14 & 23 & 31 & 37 & 45 & 54 & 77 \\
       7 & 1 & 15 & 25 & 31 & 37 & 47 & 53 & 78 \\
       8 & 4 & 17 & 28 & 37 & 40 & 48 & 57 & 80 \\
      \hline
      \end{tabular}
      \end{center}
  \caption{InferNet top-$N$ target category accuracy across error fixations (rows) where $N = 1, 2, ..., 128$ (columns)}\label{cateevalall}
\end{table}

\begin{table}
\centering
\scriptsize
\begin{tabular}{c|cc|cccccccc}

                     & \multicolumn{2}{c|}{Object Arrays} & \multicolumn{4}{c}{Natural Images}                   \\ \hline
\#Error Fix.           & 1        & 3       & 1      & 3       & 5       & 7    \\ \hline
InferNet               & 6.87     & 3.10    & 12.83  & 22.48   & 24.35   & 28.28 \\ \hline
Layer 5                & 1.88     & 1.58    & 9.35	 & 	17.11  & 17.24   & 20.91 \\
Layer 10               & 3.98     & 0.67    & 14.69	 &24.82	   &	25.16&	26.97\\
Layer 17               & 5.96     & 1.99    & 16.50  &	19.28  &	22.42&	26.43\\
Layer 23               & 7.46     & 0.01    & 13.32  &	24.72  &	28.07&	23.56\\
Layer 24               & 6.60     & 3.28    & 18.53  &	28.04  &	30.59&	30.42\\
Layer 30               & 8.21     & 3.08    & -      &	4.45   &	6.03 &	3.36\\
Layer 31               & 7.56     & 2.34    & -      &	6.15   &	5.00 &	3.93\\ \hline
Max + Max              & 6.87     & 1.13    & 12.84  &	21.11  &	22.96&	24.49\\
Mean + Max             & 7.01     & 2.63    & 8.67   &	11.97  &	14.22&	16.05\\
Mean + Mean            & 7.01     & 3.68    & 8.67   &	9.68   &	10.61&	13.31\\ \hline
Humans                 & -        & -       & 60.87  & 67.33   & 38.18   & 35.65 \\
Common Fix.            & 10.96     & 1.52   & 20.09	 &30.35    & 24.98   & 25.77\\ \hline

\end{tabular}
\caption{Target inference relative performance (\%) of ablated models compared with chance $T$ error fixations (the larger, the better). (-) means not significantly better than chance. Layer number refers to the index in the VGG16 network \cite{Simonyan14c}. The first row (InferNet) corresponds to the full model, whereas the other rows show the predictions using either only one feature similarity map from $M_{i1}$ to $M_{i7}$ in \textbf{Fig.~\ref{fig:model}} or their combinations.
}\vspace{-1mm}
\label{tab:ablationstudy}
\end{table}

\subsection{Target category inference}

In addition to inferring the target location, we asked whether InferNet could infer the categorical content of what the subjects were looking for. We selected 100 images (out of 240) where the target categories were in ImageNet. InferNet predicted the probability that the target belonged to each one of 1,000 categories by leveraging on the weights pre-trained on ImageNet and accumulating those probabilities across error fixations. Given even only one error fixation, InferNet surpassed the chance model ($1/1000$) (\textbf{Table~\ref{cateevalall}}). As expected, the target category inference accuracy increased with $T$. 
Given 8 error fixations, InferNet correctly inferred the target category with top-4 accuracy of $28\%$ out of 1,000 categories.
These results show that InferNet can not only infer location goals but also categorical content goal information.

\subsection{Ablation effect on target inference}

To evaluate the contribution of different layers, we tested each feature similarity map $M_{j}$ (\textbf{Table~\ref{tab:ablationstudy}}). Target inference was better with higher layer feature similarity maps.


Averaging instead of max-pooling the likelihood maps did not yield any significant improvements in object arrays. However, alternative ways of combining feature similarity maps led to large differences in natural images. InferNet outperforms the alternative models, suggesting that error fixations are guided by sub-patterns of the search target \cite{rajashekar2006visual}.

InferNet disregards error fixation order. Incorporating fixation order did not impact performance.

We asked how well humans can infer another subject's intentions by conducting additional psychophysics experiments (\textbf{Table~\ref{tab:ablationstudy}}). Subjects were not able to solve the inference problem in object arrays but were better than InferNet in natural images, perhaps by using contextual cues \cite{zhang2020cvpr}. To investigate the between-subject variability, we created a new model using only fixations that are common across subjects. The results (last row) show that in some (but not all) cases, InferNet can overcome the consequences of variability in human scanpath patterns. 


\section{Conclusion}
We proposed a computational model (InferNet) to infer intentions from observed eye movement behaviors. We evaluated the model in two visual search tasks. InferNet can predict the subjects' goals (what they are looking for), in object array images and in natural images, by using the set of non-target error fixations, and without any domain-specific training. 
This proof-of-principle demonstration provides a possible solution to effectively estimate what the subject is searching for in complex images. The results suggest that computational models can make reasonable conjectures to predict human intentions from behavioral data.


{\scriptsize
\bibliographystyle{ieee_fullname}
\bibliography{Mengmibib}

\begin{thebibliography}{10}\itemsep=-1pt

\bibitem{betz2010investigating}
Betz et al.
\newblock Investigating task-dependent top-down effects on overt visual
  attention.
\newblock {\em JoV}, 10(3):15--15, 2010.

\bibitem{borji2015eyes}
Borji et al.
\newblock What do eyes reveal about the mind?: Algorithmic inference of search
  targets from fixations.
\newblock {\em Neurocomputing}, 149:788--799, 2015.

\bibitem{castelhano2009viewing}
Castelhano et al.
\newblock Viewing task influences eye movement control during active scene
  perception.
\newblock {\em Journal of vision}, 9(3):6--6, 2009.

\bibitem{eckstein2007similar}
Eckstein et al.
\newblock Similar neural representations of the target for saccades and
  perception during search.
\newblock {\em Journal of Neuroscience}, 27(6):1266--1270, 2007.

\bibitem{henderson2013predicting}
Henderson et al.
\newblock Predicting cognitive state from eye movements.
\newblock {\em PloS one}, 8(5):e64937, 2013.

\bibitem{itti1998model}
Itti et al.
\newblock A model of saliency-based visual attention for rapid scene analysis.
\newblock {\em TPAMI}, 20(11):1254--1259, 1998.

\bibitem{iqbal2004using}
Iqbal et al.
\newblock Using eye gaze patterns to identify user tasks.
\newblock In {\em The Grace Hopper Celebration of Women in Computing}, pages
  5--10, 2004.

\bibitem{rajashekar2006visual}
Rajashekar et al.
\newblock Visual search: Revealing the influence of structural cues by
  gaze-contingent classification image analysis.
\newblock {\em JoV}, 6(4):7--7, 2006.

\bibitem{alexander2011visual}
Robert~G et al.
\newblock Visual similarity effects in categorical search.
\newblock {\em Journal of Vision}, 11(8):9--9, 2011.

\bibitem{Simonyan14c}
Simonyan et al.
\newblock Very deep cnn for large-scale image recognition.
\newblock {\em CoRR}, abs/1409.1556, 2014.

\bibitem{wolfe2007guided}
Wolfe et al.
\newblock Guided search 4.0.
\newblock {\em Integrated models of cognitive systems}, pages 99--119, 2007.

\bibitem{zelinsky2013eye}
Zelinsky et al.
\newblock Eye can read your mind: Decoding gaze fixations to reveal categorical
  search targets.
\newblock {\em JoV}, 13(14):10--10, 2013.

\bibitem{zhang2018finding}
Zhang et al.
\newblock Finding any waldo with zero-shot invariant and efficient visual
  search.
\newblock {\em NC}, 9(1):1--15, 2018.

\bibitem{zhang2020cvpr}
Zhang et al.
\newblock Putting visual object recognition in context.
\newblock {\em arXiv}, 1911.07349, 2020.

\end{thebibliography}
}

\end{document}


%
\title{Supplementary Material for \\ What am I searching for?}
\author{Anonymous}
\maketitle
\section{Layer specification for InferNet}
In InferNet, the following specific layers were selected: $M_{i1}$ (layer $j=5$ of VGG16, size $101 \times 101$), $M_{i2}$ (layer $j=10$, $52 \times 52$), $M_{i3}$ (layer $j=17$, $27 \times 27$), $M_{i4}$ (layer $j=23$, $27 \times 27$), $M_{i5}$ (layer $j=24$, $15 \times 15$), $M_{i6}$ (layer $j = 30$, $15 \times 15$), $M_{i7}$ (layer $j=31$, $9 \times 9$). The layer number refers to the layer index in VGG16 \cite{Simonyan14c}.

\section{Low-level feature normalization in object array dataset}
We selected segmented objects without occlusion from natural images in MSCOCO dataset \cite{lin2014microsoft} from 6 categories: sheep, cattle, cats, horses, teddy bears and kites. Due to the uncontrolled and diverse nature of these stimuli, they may differ in low-level properties that could contribute to visual search performance. To minimize such contributions, we took the following steps: (1) we normalized the object areas within a bounding box ($156 \times 156$ pixels) while maintaining their aspect ratios; (2) we converted them to greyscale images; (3) we equalized their luminance histograms, and (4) we randomly rotated the objects in 2D. We evaluated the output of these steps by building a 6-category classifier based on several models of simple features and verified that all of these low-level models were only marginally above chance levels.

\section{Role of fixation duration and sequence on target inference performance}
We studied the role of the locations and the sequence order of error fixations in target inference. We created the ablated model and trained it using supervised learning to predict the final inference map directly: (1) generate a binary error fixation map masked with gaussian kernels to denote the locations of error fixations and the magnitude of gaussian kernels vary depending on the fixation order. The higher intensity of the gaussian mask is applied at the error fixation, the more recent the error fixation is. (2) concatenate this fixation map with the search image as inputs to a feed-forward 2D convolution neural network. (3) Kullback?Leibler divergence loss is computed between the predicted inference map and the ground truth map where 1 denotes the target location and 0 otherwise.

\begin{table*}
\centering
\caption{Target inference performance of ablated models in two datasets: object arrays and natural images. Based on the number of error fixations given, we report the relative performance (\%) compared with the chance model (the larger, the better) after taking into account of error fixation locations and fixation order information.\vspace{-1mm}}\vspace{-1mm}
\label{tab:ablationstudy}
\scriptsize
\begin{tabular}{c|cccc|cccccccc}
                       & \multicolumn{4}{c|}{Object Arrays} & \multicolumn{8}{c}{Natural Images}                   \\ \hline
\# Error Fixation & 1       & 2      & 3      & 4      & 1    & 2    & 3    & 4    & 5    & 6    & 7    & 8    \\ \hline
InferNet (our model)   & 6.87     & 5.25    & 3.10    & -    & 12.83 & 19.67 & 22.48   & 24.20 & 24.35 & 25.59 & 28.28 & 18.14   \\ \hline
Location + Sequence    & 7.14     & 10.61    & 7.87    & 0.78    & -37.78 & -26.14 & -30.16 & -31.91 & -28.65 & -34.25 & -30.92 & -34.81 \\
\end{tabular}
\end{table*}

In Table~\ref{tab:ablationstudy}, we reported the result (Location+Sequence) in both object arrays and natural images. This ablated model which takes location and order information into account performs equally well as InferNet in object arrays but much worse than InferNet in natural images. It is surprising that this experimental result seems to suggest the location and order information of error fixations do not matter much in target inference task; however, a more detailed analysis is needed to validate this point in the future.

\section{Feature similarity analysis}

\begin{figure*}[t]
\centering
    \subfloat[Last fixations on target objects]{\includegraphics[width= 6.3cm]{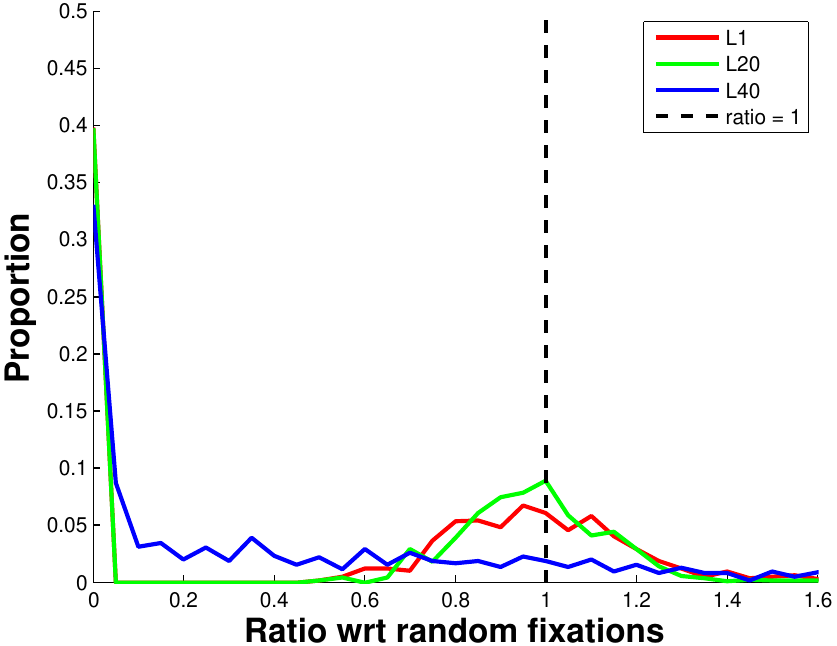}\label{fig:dlast}}\hspace{0.2cm}
    \subfloat[Error fixations on non-target objects]{\includegraphics[width= 6.3cm]{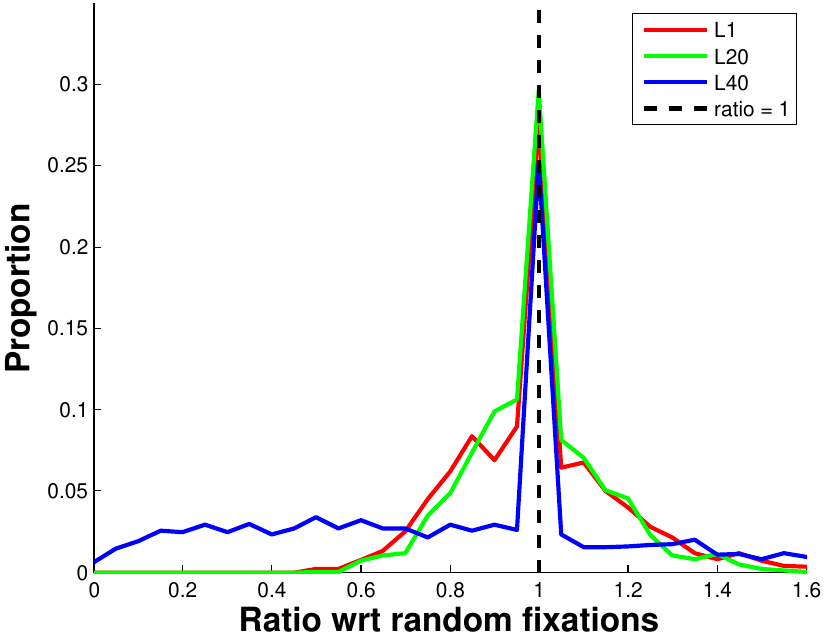}\label{fig:derror}}
    \caption{Feature similarity analysis between error fixations or last on-target fixations and the given target exemplars across layers in object arrays in human visual search tasks. We showed the distribution of the ratio between L2 distance of last on-target fixation-target pair versus random fixation-target pairs in subplot (a) and error fixation-target pair versus random fixation-target pair in subplot (b) using features extracted from different layers in pre-trained VGG16. Similar trends are observed across all layers. For simplicity, we only presented results on Layer 1 (red), Layer 20 (green) and Layer 40 (blue). The dash line denotes the ratio equal to 1. If there are more error fixations-target pairs which are more similar than random fixation-target pairs, the distribution of the ratios is skewed to the left of dash line (ratio $=$ 1) and vice versa. \vspace{-2mm}}\vspace{-2mm}\label{featuresimilarityanalysis}
\end{figure*}

According to \cite{eckstein2007similar,alexander2011visual,wolfe2007guided}, the error fixations share more target-similar features than distractors. Thanks to the rich deep features from VGG16 pre-trained on object classification task, we provided a detailed analysis of the feature similarity between pairs of error fixations-target and pairs of random fixations-target in object arrays within layers and across layers in VGG16.

Figure~\ref{fig:derror} shows the distribution of the ratio between L2 distance of error fixation-target pairs versus random fixation-target pairs for layer 1, 20, 40. If error fixations share more target-similar features, the distance between error fixations and the targets is expected to be smaller than random fixation-target pairs which makes the ratio less than 1. We observed that the ratio distribution is skewed to the left of dash line (ratio equals 1) within layers which validates the point that there are more error fixations which share more similarities with the targets than random fixations versus the targets. The ratio distribution in higher layers (from 1 to 40) becomes more dispersed with larger variance which suggests features in higher layers have more discriminative power to distinguish target-similar objects among other distractors. The area under the curve to the left of the dash line becomes more dominant than the other half as the layer number increases.

For comparison purposes, we also provided the ratio distribution between last on-target fixation-target pairs versus random fixation-target pairs (See Figure~\ref{fig:dlast}). As expected, since the last fixations are on the target, they should share more feature similarities than random fixation-target pairs. Thus, most of the pairs should have ratio less than 1. Across all layers, we observed that the area under the curve for ratio $<1$ is far greater than the other half. As the layer number increases, the area under the curve on the left of the dash line (ratio = 1) increases.


{\small
\bibliographystyle{aaai}
\bibliography{Mengmibib}
}